\documentclass[runningheads]{llncs}
\usepackage[hyphens]{url}
\usepackage{hyperref}           
\usepackage{booktabs}       
\usepackage{amsfonts}       
\usepackage{nicefrac}       
\usepackage{microtype}      
\usepackage{multirow}
\usepackage{graphicx}
\usepackage[labelfont=bf]{caption}
\usepackage{subcaption}
\usepackage{multicol}
\usepackage{footmisc}
\usepackage{ctable}
\usepackage{amsmath}
\usepackage{newtxtext,newtxmath}

\begin{document}
\mainmatter              

\newcommand{\workshopdata}{\texttt{FakeNews-19}}
\newcommand{\yejindata}{\texttt{Tweets-19}}

\title{Model Generalization on COVID-19 Fake News Detection}
%
%


\authorrunning{Bang et al.} 

\author{Yejin Bang\thanks{\hspace{1mm}These authors contributed equally.} \and Etsuko Ishii$^*$\and Samuel Cahyawijaya$^*$\and Ziwei Ji$^*$ \and Pascale Fung}

\institute{Center for Artificial Intelligence Research (CAiRE)\\
Department of Electronic and Computer Engineering\\
The Hong Kong University of Science and Technology, Clear Water Bay, Hong Kong\\
\email{\{yjbang,eishii,scahyawijaya,zjiad\}@connect.ust.hk}}

\maketitle              

\begin{abstract}
Amid the pandemic COVID-19, the world is facing unprecedented \textit{infodemic} with the proliferation of both fake and real information. Considering the problematic consequences that the COVID-19 fake-news have brought, the scientific community has put effort to tackle it. To contribute to this fight against the infodemic, we aim to achieve a robust model for the COVID-19 fake-news detection task proposed at CONSTRAINT 2021 (\workshopdata) by taking two separate approaches: 1) fine-tuning transformers based language models with robust loss functions and 2) removing harmful training instances through influence calculation. We further evaluate the robustness of our models by evaluating on different COVID-19 misinformation test set (\yejindata) to understand model generalization ability. With the first approach, we achieve 98.13\% for weighted F1 score (W-F1) for the shared task, whereas 38.18\% W-F1 on the \yejindata~ highest. On the contrary, by performing influence data cleansing, our model with 99\% cleansing percentage can achieve 54.33\% W-F1 score on \yejindata~with a trade-off. By evaluating our models on two COVID-19 fake-news test sets, we suggest the importance of model generalization ability in this task to step forward to tackle the COVID-19 fake-news problem in online social media platforms.

\end{abstract}
%

 
\section{Introduction}
As the whole world is going through a tough time due to the pandemic COVID-19, the information about COVID-19 online grew exponentially. It is the first global pandemic with the 4th industrial revolution, which led to the rapid spread of information through various online platforms. It came along with \textit{Infodemic}. The infodemic results in serious problems that even affects people's lives, for instance, a fake news ``Drinking bleach can cure coronavirus disease'' led people to death\footnote{\url{https://www.bbc.com/news/world-53755067}}. Not only the physical health is threatened due to the fake-news, but the easily spread fake-news even affects the mental health of the public with restless anxiety or fear induced by the misinformation~\cite{xiong2020impact}.

With the urgent calls to combat the infodemic, the scientific community has produced intensive research and applications for analyzing contents, source, propagators, and
propagation of the misinformation~\cite{pennycook2020fighting,mian2020coronavirus,kouzy2020coronavirus,brennen2020types,lee2020misinformation} and providing accurate information through various user-friendly platforms~\cite{li2020jennifer,su2020caire}. The early published fact sheet about the COVID-19 misinformation suggested 59\% of the sampled pandemic-related Twitter posts are evaluated as fake-news~\cite{brennen2020types}. To address this, a huge amount of tweets is collected to disseminate the misinformation~\cite{medford2020infodemic,mourad2020critical,shahi2020exploratory,alam2020fighting}. Understanding the problematic consequences of the fake-news, the online platform providers have started flag COVID-19 related information with an ``alert'' so the audience could be aware of the content. However, the massive amount of information flooding the internet on daily basis makes it challenging for human fact-checkers to keep up with the speed of information proliferation \cite{shao2018anatomy}. The automatic way to aid the human fact-checker is in need, not just for COVID-19 but also for any infodemic that could happen unexpectedly in the future.

In this work, we aim to achieve a robust model for the COVID-19 fake-news detection shared task proposed by Patwa.~et al.~\cite{patwa2020fighting} with two approaches 1) fine-tuning classifiers with robust loss functions and 2) removing harmful training instances through influence calculation. We also further evaluate the adaptability of our method out of the shared task domain through evaluations on different COVID-19 misinformation tweet test set~\cite{alam2020fighting}. We show a robust model with high performance over two different test sets to step forward to tackle the COVID-19 fake-news problem in social media platforms.

\begin{table}[t]
  \caption{Dataset Statistics.}
  \centering
  \resizebox{0.42\linewidth}{!}{
\begin{tabular}{cccccc}
\toprule
 & \multicolumn{3}{c}{\workshopdata} & \multicolumn{2}{c}{\yejindata} \\ \cmidrule(lr){2-4}\cmidrule(lr){5-6} 
Label \hspace{1pt} & \hspace{1pt} Train & Valid & Test \hspace{1pt} & \hspace{1pt} Valid &  Test \\ \midrule
Real & 3360 & 1120 & 1120 & 51 & 172 \\
Fake & 3060 & 1020 & 1020 & 9 & 28  \\ \midrule
Total & 6420 & 2140 & 2140 & 60 & 200  \\ \bottomrule
\end{tabular}}
\label{table:data_statistics}
\end{table}

\section{Dataset}
\textbf{Fake-News COVID-19 (\workshopdata)} A dataset released for the shared task of CONSTRAINT 2021 workshop ~\cite{patwa2021overview}, which aims to combat the infodemic regarding COVID-19 across social media platforms such as Twitter, Facebook, Instagram, and any other popular press releases. The dataset consists of 10,700 social media posts and articles of real and fake news, all in English. The details of the statistic are listed in Table~\ref{table:data_statistics}. Each social media post is manually annotated either as ``Fake'' or ``Real'', depending on its veracity.

\begin{table}[t]
  \caption{Top-10 most frequent words on \workshopdata~and \yejindata}
  \centering
  \resizebox{\linewidth}{!}{
    \begin{tabular}{ccccl}
    \toprule
    Dataset & & Label & & Most frequent words \\ 
    \midrule\midrule
    \multirow[c]{2}{*}{Real} & & \workshopdata & & cases, \#covid19, new, covid, tests, people, states, deaths, total, testing \\
    \cmidrule(lr){2-5}
    & & \yejindata & & \#coronavirus, covid, cases, \#covid19, people, virus, corona, health, spread, us \\
    \midrule 
    \multirow[c]{2}{*}{Fake} & & \workshopdata & & covid, coronavirus, people, virus, vaccine, \#coronavirus, trump, says, new, \#covid19 \\
    \cmidrule(lr){2-5}
     & & \yejindata & & virus, corona, coronavirus, covid, \#coronavirus, fake, news, get, really, media \\
    \bottomrule
    \end{tabular}
  }
\label{table:frequest_word_statistics}
\end{table}
\noindent\textbf{Tweets COVID-19 (\yejindata)} To evaluate the generalizability of trained models test setting, we take the test set from \cite{alam2020fighting}, which is also released for fighting for the COVID-19 Infodemic tweets. The tweets are annotated with fine-grained labels related to disinformation about COVID-19, depending on the interest of different parties involved in the Infodemic. We took the second question, ``\textit{To what extent does the tweet appear to contain false information?}'', to incorporate with our binary setting. Originally, it is answered in five labels based on the degree of the falseness of the tweet. Instead of using the multi-labels, we follow the binary setting as the data releaser did to map to ``Real'' and ``Fake'' labels for our experiments. For our cleansing experiment, we split the dataset into validation and test set with equal label distribution. The detail is listed in Table \ref{table:data_statistics}. The most frequent words after removing stopwords on each dataset is listed in Table \ref{table:frequest_word_statistics}.

\section{Methodology}

\subsection{Task and Objective}
The main task is a binary classification to determine the veracity for the given piece of text from social media platforms and assign the label either ``Fake'' or ``Real''. We aim to achieve a robust model in this task with a consideration on both high performance on predicting labels on \workshopdata~shared task and generalization ability through performance on \yejindata~with two separate approaches described in the following Sections (\ref{subsection:approach1} and  \ref{subsection:approach2}). Note that models are trained only with \workshopdata~train set.

\subsection{Approach 1: Fine-tuning Pre-trained Transformer based Language Models with Robust Loss Functions}
\label{subsection:approach1}

When handling text data, Transformers~\cite{vaswani2017attention} based language models (LM) are commonly used as feature extractors~\cite{devlin-etal-2019-bert,Lan2020ALBERT,liu2019roberta} thanks to publicly released large-scale pre-trained language models (LMs). We adopt different Transformer LMs with a feed-forward classifier trained on top of each model. The list and details of models are described in Section~\ref{experiment1}.
As reported in~\cite{xia2020partdependent,karimi2020deep,kumar2018robust}, robust loss functions help to improve the deep neural network performance especially with noisy datasets constructed from social medium. In addition to the standard cross-entropy loss (CE), we explore the following robust loss functions: symmetric cross-entropy (SCE)~\cite{wang2019symmetric}, the generalized cross-entropy (GCE)~\cite{zhang2018generalized}, and curriculum loss (CL)~\cite{lyu2019curriculum}. Inspired by the symmetric Kullback-Leibler divergence, SCE takes an additional term called reverse cross-entropy to enhance CE symmetricity. GCE takes the advantages of both mean absolute error being noise-robust and CE performing well with challenging datasets. CL is a recently proposed 0-1 loss function which is a tighter upper bound compared with conventional summation based surrogate losses, which follows the investigation of 0-1 loss being robust~\cite{hu2018does}.

\subsection{Approach 2: Data Noise Cleansing based on Training Instance Influence}
\label{subsection:approach2}
This approach is inspired by the work of Kobayashi et al.~\cite{kobayashi-etal-2020-efficient}, which proposes an efficient method to estimate the influence of training instances given a target instance by introducing \textit{turn-over dropout} mechanism. We define $D^{\mathrm{trn}}$ $= \{d_1^{\mathrm{trn}}, d_2^{\mathrm{trn}}, \dots, d_k^{\mathrm{trn}}\}$ as a training dataset with $k$ training sample and $\mathcal{L}(f,d)$ as a loss function calculated from a model $f$ and a labelled sample $d$. In turn-over dropout, a specific dropout mask $m_i \in \{0, \frac{1}{p}\}$ with dropout probability $p$ is applied during training to zeroed-out a set of parameters $\theta \in \mathbb{R}^n$ from the model $f$ for each training instance $d_i^{\mathrm{trn}}$. With this approach, every single sample in the training set is trained on a unique sub-network of the model. 

We define $h(d_i^{\mathrm{trn}})$ is a function to map a training data $d_i^{\mathrm{trn}}$ into the specific mask $m_i$. The influence score $I(d^{\mathrm{tgt}}, d_i^{\mathrm{trn}}, f)$ for each target sample $d^{\mathrm{tgt}}$ is defined as follow:

\begin{equation*}
I(d^{\mathrm{tgt}}, d_i^{\mathrm{trn}}, f) = \mathcal{L}(f^{\widetilde{h(d_i^{\mathrm{trn}})}}, d^{\mathrm{tgt}}) - \mathcal{L}(f^{h(d_i^{\mathrm{trn}})}, d^{\mathrm{tgt}}),
\end{equation*}

\noindent where $\widetilde{m_i}$ is the flipped mask of the original mask $m_i$, i.e., $\widetilde{m_i} = \frac{1}{p} - m_i$, and $f^{m_i}$ is the sub-network of the model with the mask $m_i$ applied. Intuitively, the influence score indicates the contribution of a training instance $d_i^{\mathrm{trn}}$ to the target instance $d^{\mathrm{tgt}}$. A positive influence score indicates $d_i^{\mathrm{trn}}$ reduces the loss of $d^{\mathrm{tgt}}$ and a negative influence score indicates $d_i^{\mathrm{trn}}$ increases the loss of $d^{\mathrm{tgt}}$, and the magnitude of the score indicates how strong the influence is. To calculate the total influence score of a training data $d_i^{\mathrm{trn}}$ over multiple samples from a given target set $D^{\mathrm{tgt}} = \{d_1^{\mathrm{tgt}}, d_2^{\mathrm{tgt}}, \dots, d_k^{\mathrm{tgt}}\}$, we accumulate each individual influence score by:

\begin{equation*}
I_{\mathrm{tot}}(D^{\mathrm{tgt}}, d_i^{\mathrm{trn}}, f) = \sum_{j=1}^K I(d_j^{\mathrm{tgt}}, d_i^{\mathrm{trn}}, f).
\end{equation*}

The total influence score $I_{\mathrm{tot}}$ can be used to remove harmful instances, which only add noise or hinder generalization of the model, from the training set by removing top-$n$\% of training instances with the smallest total influence score from the training data. We refer to our data cleansing method as influence-based cleansing which can remove noisy data and further improve model robustness and adaptability.

\section{Experiment 1: Fine-tuning LMs with Robust Loss Functions}

\begin{table}[t]
  \caption{Results on \workshopdata~test set using large language models. Underline indicates the best performance on each model. Acc. and W-F1 stands for Accuracy and weighted F1 respectively. SVM is placed under the column of CE for ease of comparison.}
  \label{table:result_largemodels}
\centering
\begin{tabular}{ccccccccc}

\toprule
Loss Functions & \multicolumn{2}{c}{CE} & \multicolumn{2}{c}{SCE} & \multicolumn{2}{c}{GCE} & \multicolumn{2}{c}{CL} \\ \cmidrule(lr){2-3}\cmidrule(lr){4-5}\cmidrule(lr){6-7}\cmidrule(lr){8-9}
Models & Acc. & W-F1 & Acc. & W-F1 & Acc. & W-F1 & Acc. & W-F1 \\\midrule\midrule
TF-IDF SVM~\cite{patwa2020fighting} & 93.32 & 93.32 & - & - & - & - & - & - \\
ALBERT-base & \underline{97.34}	& \underline{97.33} & 96.82 & 96.82 & 96.45 & 96.44 & 96.73 & 96.72 \\ 
BERT-base & \underline{97.99} & \underline{97.99} & 97.15 & 97.14 & 97.66 & 97.66 & 97.71 & 97.7 \\
BERT-large & 97.15 & 97.14 & 96.92 & 96.91 & \underline{97.29} &  \underline{97.28} & 97.24 & 97.23 \\ 
RoBERTa-base & \underline{97.94} & \underline{97.94} & 97.52 & 97.51 & 97.57 & 97.56 & 97.62 & 97.61 \\
RoBERTa-large & \textbf{\underline{98.13}} & \textbf{\underline{98.13}} & 97.90 & 97.89 & 97.48 & 97.47 & 97.48 & 97.47 \\
\bottomrule
\end{tabular}
\end{table}

\subsection{Experiment Set-up}
\label{experiment1}
We set up the baseline of our experiment from \cite{patwa2020fighting}, an SVM model trained with features extracted from extracted by using TF-IDF. We try five different pre-trained BERT-based models, including ALBERT-base~\cite{Lan2020ALBERT}, BERT-base, BERT-large~\cite{devlin-etal-2019-bert}, RoBERTa-base, and RoBERTa-large~\cite{liu2019roberta}. We fine-tune the models on \workshopdata~train set with the classification layers on the top exploiting the pre-trained models provided by~\cite{wolf-etal-2020-transformers}. We train each model with four different loss functions, which are CE, SCE, GCE, and CL. The hyperparameters are searched with learning rate of $1\mathrm{e}{-6}, 3\mathrm{e}{-6}, 5\mathrm{e}{-6}$ and epoch of $1,3,5,10$ and the best combination is chosen based on performance on \workshopdata~validation set. The robustness of fine-tune models is then evaluated on both \workshopdata~and \yejindata~test sets. In this experiment, we mainly focus our evaluation on the Weighted-F1 (W-F1) score.
\subsection{Experimental Results}
Table~\ref{table:result_largemodels} reports the result of on \workshopdata~task. Across all settings, RoBERTa-large trained with CE loss function achieved the highest W-F1 scores, 98.13\%, with a gain of 4.81\% in W-F1 compared to the TF-IDF SVM baseline. Except for BERT-large, all other models achieved their best performance when fine-tuned with CE loss function. The robust loss functions did not contribute in terms of improving the performance of predicting the labels. In other words, the large-scale LMs could extract high-quality features that the noise with \workshopdata~was barely available for the robust loss functions to contribute. 

In Table~\ref{table:result_approach1_yejin}, we show the inference results on \yejindata; unlike the successful result on \workshopdata~RoBERTa-large with CE scores only 33.65\% of W-F1 on \yejindata, showing that the generalization of the model is not successful. Instead, the highest performance could be achieved with BERT-large with SCE with 38.18\%, which is 4.53\% gain compared to RoBERTa-large with CE. Interestingly, across all models, the highest performance when fine-tuned with the robust loss functions, SCE, GCE, and CL. This shows the robust loss functions help to improve the generalization ability of models. For instance, the RoBERTa-large could gain 3.85\% with CL loss function, compared to its performance with CE. Considering that RoBERTa-large with CL achieves 97.47\%, which is only 0.66\% loss from the highest performance, it can be considered as a fair trade-off for selecting RoBERTa-large with CL could as a robust model, which achieves high performance on \workshopdata~as well as generalizes better on \yejindata.

Overall, while LMs with robust loss functions could achieve the highest 98.13\% and lowest 96.44\% on \workshopdata, performance on \yejindata~is comparatively poor as lower than 40\% and even results in 22.85\% lowest for W-F1. It could be inferred that the test set distributions are distinct although they are both related to COVID-19 infodemic and share the same data source, Twitter. This could be explained that CL is more robust to noisy labels, where \workshopdata~labels are considered to be noisy to \yejindata~test set. Further analysis is in Section~\ref{subsection:data_distribution}.

\begin{table}[t]
  \caption{Results on \yejindata~test set of large language model classifiers. Underlined results indicate the highest performance within each model.}
  \label{table:result_approach1_yejin}
\centering
\begin{tabular}{ccccccccc}
\toprule
Loss Functions & \multicolumn{2}{c}{CE} & \multicolumn{2}{c}{SCE} & \multicolumn{2}{c}{GCE} & \multicolumn{2}{c}{CL} \\ \cmidrule(lr){2-3}\cmidrule(lr){4-5}\cmidrule(lr){6-7}\cmidrule(lr){8-9}
Models & \hspace{3pt} Acc. & W-F1 \hspace{3pt} & \hspace{3pt} Acc. & W-F1 \hspace{4pt} & \hspace{3pt} Acc. & W-F1 \hspace{3pt} & \hspace{3pt} Acc. & W-F1 \\\midrule\midrule
ALBERT-base & 35.38 & 35.07 & 36.15 & 35.69 & \underline{37.69} & \underline{37.16} & 33.85 & 33.59 \\ 
BERT-base & 23.08 & 22.85 & \underline{33.08} & \underline{32.93} & 31.15 & 31.10 & 24.62 & 24.50 \\
BERT-large & 32.69 & 32.57 & \textbf{\underline{38.85}} & \textbf{\underline{38.18}} & 32.69 & 32.57 & 31.54 & 31.47 \\ 
RoBERTa-base & 28.08 & 28.08 & \underline{36.92} & \underline{36.38} & 33.46 & 33.24 & 29.62 & 29.61 \\
RoBERTa-large & 33.85 & 33.65 & 31.54 & 31.47 & 31.92 & 31.84 & \underline{38.08} & \underline{37.50} \\
\bottomrule
\end{tabular}
\end{table}

\section{Experiment 2: Data Cleansing with Influence Calculation}
\label{experiment2}
\subsection{Experiment Set-up}

We first fine-tune a pre-trained RoBERTa-large model with \workshopdata~train set while applying \textit{turn-over dropout} to the weight matrix on the last affine transformation layer of the model with dropout probability of $p=0.5$. We calculate the total influence score from the resulting model to the validation sets of \workshopdata~and \yejindata. We investigate the effectiveness of our data cleansing approach by removing $n$\% of training instances with the smallest total influence score with $n = \{ 1, 25, 50, 75, 99\}$. Then, we retrain the models from the remaining training data and perform an evaluation of the retrained model. All the models are trained with Cross-Entropy loss function with a fixed learning rate of $3\mathrm{e}{-6}$. We run the model for 15 epochs with the early stopping of 3.
As the baseline, we compare our method with three different approaches: 1) pre-trained RoBERTa-large model without additional fine-tuning, 2) RoBERTa-large model fine-tuned with all training data without performing any data cleansing, and 3) model trained with random cleansing using the same cleansing percentage. We run each experiment five times with different random seeds to measure the evaluation performance statistics from each experiment. 

\subsection{Experiment Result}
\label{subsection:zero_shot_result}
Based on our experiment results in Table~\ref{table:result_approach2_indian}, our influence-based cleansing method performs best for \yejindata~when the cleansing percentage is at 99\% by only using 64 most influential training data. When cleansing percentage $\geq 25\%$, our influence-cleansed model outperforms the model without cleansing and the model with the random cleansing approach in terms of both accuracy and W-F1. The pre-trained model without fine-tuning (i.e. 0 training instance) results in 34.36\% and 46.24\% W-F1 on \workshopdata~and \yejindata~respectively. Our best model produces a significantly higher F1-score compared to the pre-trained model without fine-tuning by a large margin on both \workshopdata~and \yejindata, which means that the small set of the most influential training data helps to significantly boost the generalization ability on both datasets. Furthermore, even with a high cleansing percentage, our model can maintain high evaluation performance on the \workshopdata. Specifically, our model with a 99\% cleansing percentage can produce an evaluation
performance of 61.10\% accuracy score and 54.33\% W-F1 score on \yejindata~and 87.79\% accuracy score and 87.69\% W-F1 score on \workshopdata. With this method, we could achieve an absolute gain of 20.69 W-F1 on \yejindata, a much-improved generalization ability. Compared to the highest score achieved with using the full data for training, however, there is a trade-off with 10.44\% loss for \workshopdata. This trade-off in performances on two test sets suggests a potential for handling unseen data set during the training phase.

\begin{table}[t]
  \caption{Results on \workshopdata~test set and \yejindata~test set using Data cleansing approach. Model performance is explored when $n$\% of harmful instances are dropped from the training. We run the experiments 5 times and report the mean. The underlined value indicates a higher value for comparing Influence vs. Random for each test set and each row.}
  \label{table:result_approach2_indian}
  \centering
    \begin{tabular}{ccccccccccc}
    \toprule
    \multicolumn{2}{c}{\multirow{2}{*}{\begin{tabular}[c]{@{}c@{}}Drop of \\ Instance\end{tabular}}} & \multicolumn{1}{l}{\multirow{2}{*}{\begin{tabular}[c]{@{}l@{}}Training \\ Instance\end{tabular}}} & \multicolumn{4}{c}{\textbf{\workshopdata}} & \multicolumn{4}{c}{\textbf{\yejindata}} \\ \cmidrule(lr){4-7} \cmidrule(lr){8-11} 
    \multicolumn{2}{c}{} & \multicolumn{1}{l}{} & \multicolumn{2}{c}{Influence} & \multicolumn{2}{c}{Random} & \multicolumn{2}{c}{Influence} & \multicolumn{2}{c}{Random} \\
    \cmidrule(lr){1-2} \cmidrule(lr){3-3} \cmidrule(lr){4-5} \cmidrule(lr){6-7} \cmidrule(lr){8-9} \cmidrule(lr){10-11}
    \% & \hspace{8pt} \# \hspace{8pt} & \# & \hspace{3pt} Acc. \hspace{1pt} & W-F1 \hspace{1pt} & \hspace{1pt} Acc. \hspace{1pt} & W-F1 \hspace{5pt} & \hspace{5pt} Acc. \hspace{1pt} & W-F1 \hspace{1pt} & \hspace{1pt} Acc. \hspace{1pt} & W-F1 \\ 
    \midrule \midrule
    0\% & 0 & 6420 &\textbf{98.13} & \textbf{98.13} & 98.13 & 98.13 & 33.85 & 33.65  & 33.85 & 33.65 \\
    1\% & 64 & 6356 & \underline{97.96} & \underline{97.96} & 97.40 & 97.40 & \underline{32.00} & \underline{31.76} & \underline{30.60} & 30.39  \\
    25\% & 1605 & 4815 & \underline{97.25} & \underline{97.24} & 97.14 & 97.13 & \underline{36.70} & \underline{36.12} & 32.60 & 32.33 \\
    50\% & 3210 & 3210 & \underline{97.01} & \underline{97.00} & 88.29 & 86.38 &  \underline{37.70} & \underline{37.09} & 30.80	& 30.19 \\
    75\% & 4815 & 1605 & 96.27 & 96.26 & \underline{96.34} & \underline{96.32} & \underline{39.50} & \underline{38.62} & 38.50 & 37.58 \\
    99\% & 6356 & 64 & 87.79 & 87.69 & \underline{89.13} & \underline{89.09} & \textbf{\underline{61.10}} & \textbf{\underline{54.33}} & 48.00 & 45.45 \\ \bottomrule
    \end{tabular}
\end{table}

\section{Discussion}

\subsection{Data Distribution between different \workshopdata~and \yejindata~Test Sets}
\label{subsection:data_distribution}

\begin{figure}[t]
    \centering
    \begin{subfigure}[b]{0.4\textwidth}
         \centering
         \includegraphics[width=\textwidth]{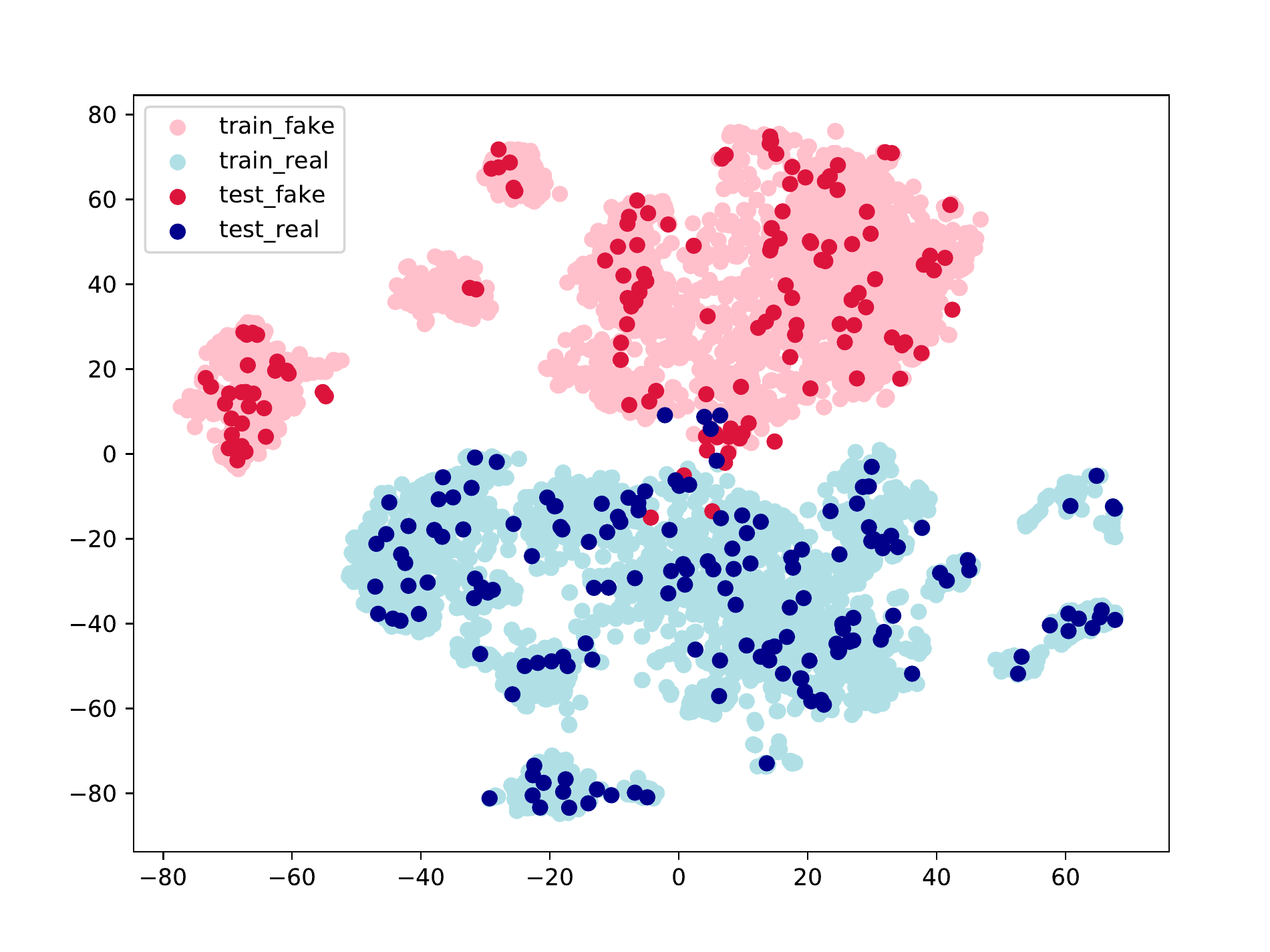}
         \caption{\workshopdata~test set.}
         \label{fig:train_valid_best}
     \end{subfigure}
     \begin{subfigure}[b]{0.4\textwidth}
         \centering
         \includegraphics[width=\textwidth]{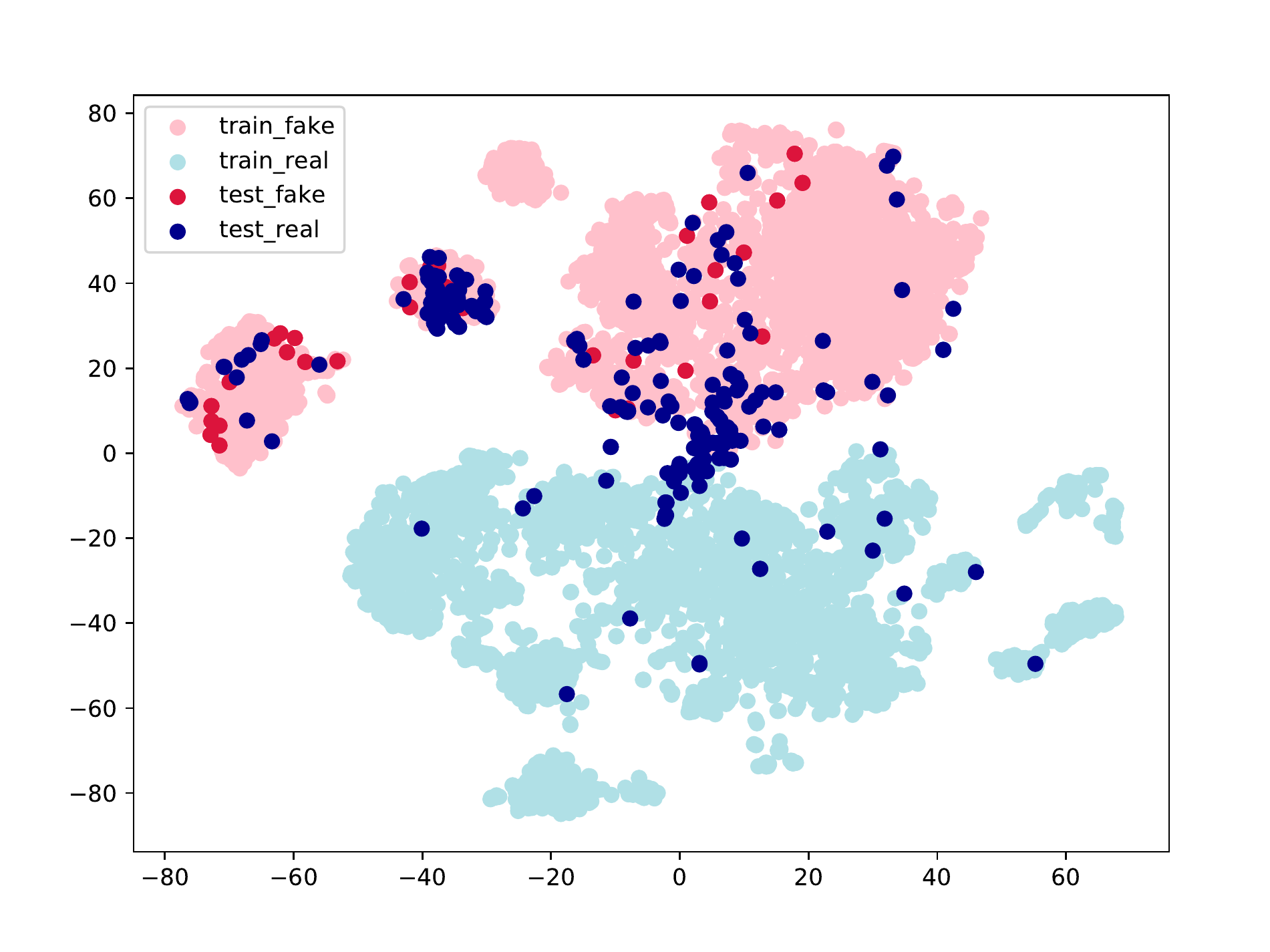}
         \caption{\yejindata~test set.}
         \label{fig:train_test_best}
     \end{subfigure}
     \caption{Datasets distribution comparison with \workshopdata~training set using t-SNE. While the distributions within \workshopdata~kept to be similar, the distribution of \yejindata~is significantly different.}
    \label{fig:tsne1}
\end{figure}

\begin{figure}[t]
    \centering
    \begin{subfigure}[b]{0.32\textwidth}
         \centering
         \includegraphics[width=\textwidth]{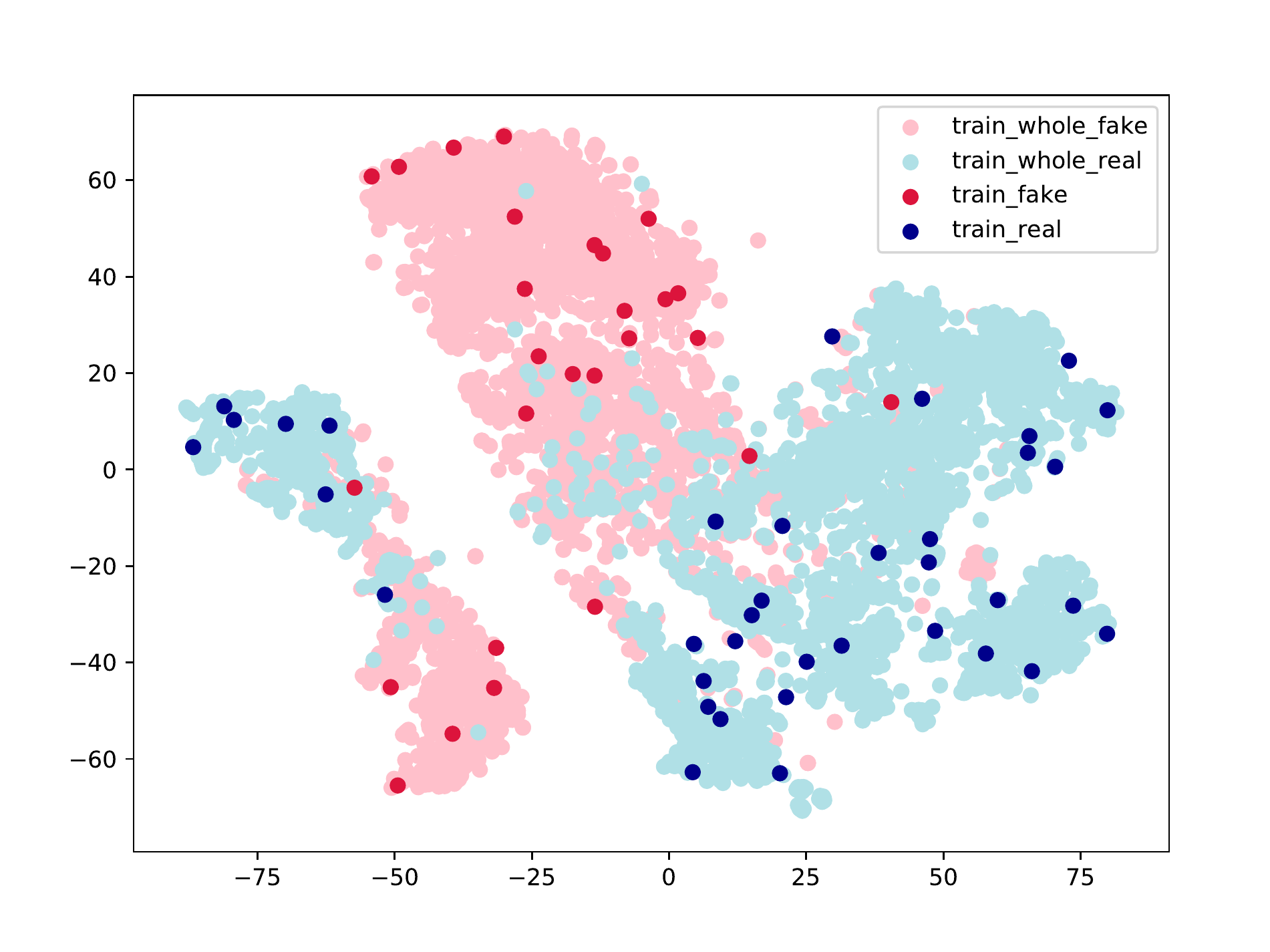}
         \caption{\workshopdata~train set.}
         \label{fig:train_clean_c99}
     \end{subfigure}
     \hfill
     \begin{subfigure}[b]{0.32\textwidth}
         \centering
         \includegraphics[width=\textwidth]{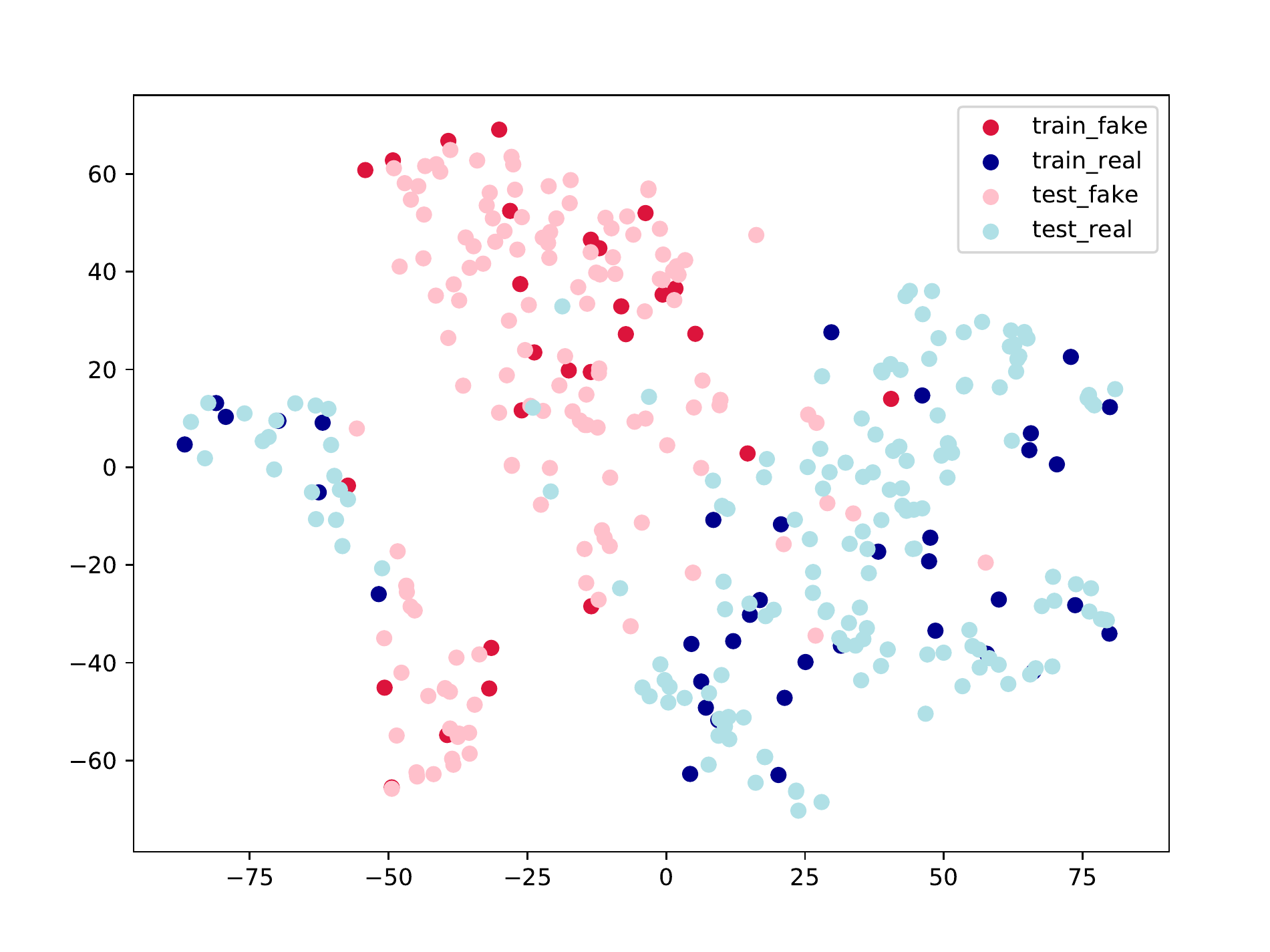}
         \caption{\workshopdata~test set.}
         \label{fig:train_valid_c99}
     \end{subfigure}
     \hfill
     \begin{subfigure}[b]{0.32\textwidth}
         \centering
         \includegraphics[width=\textwidth]{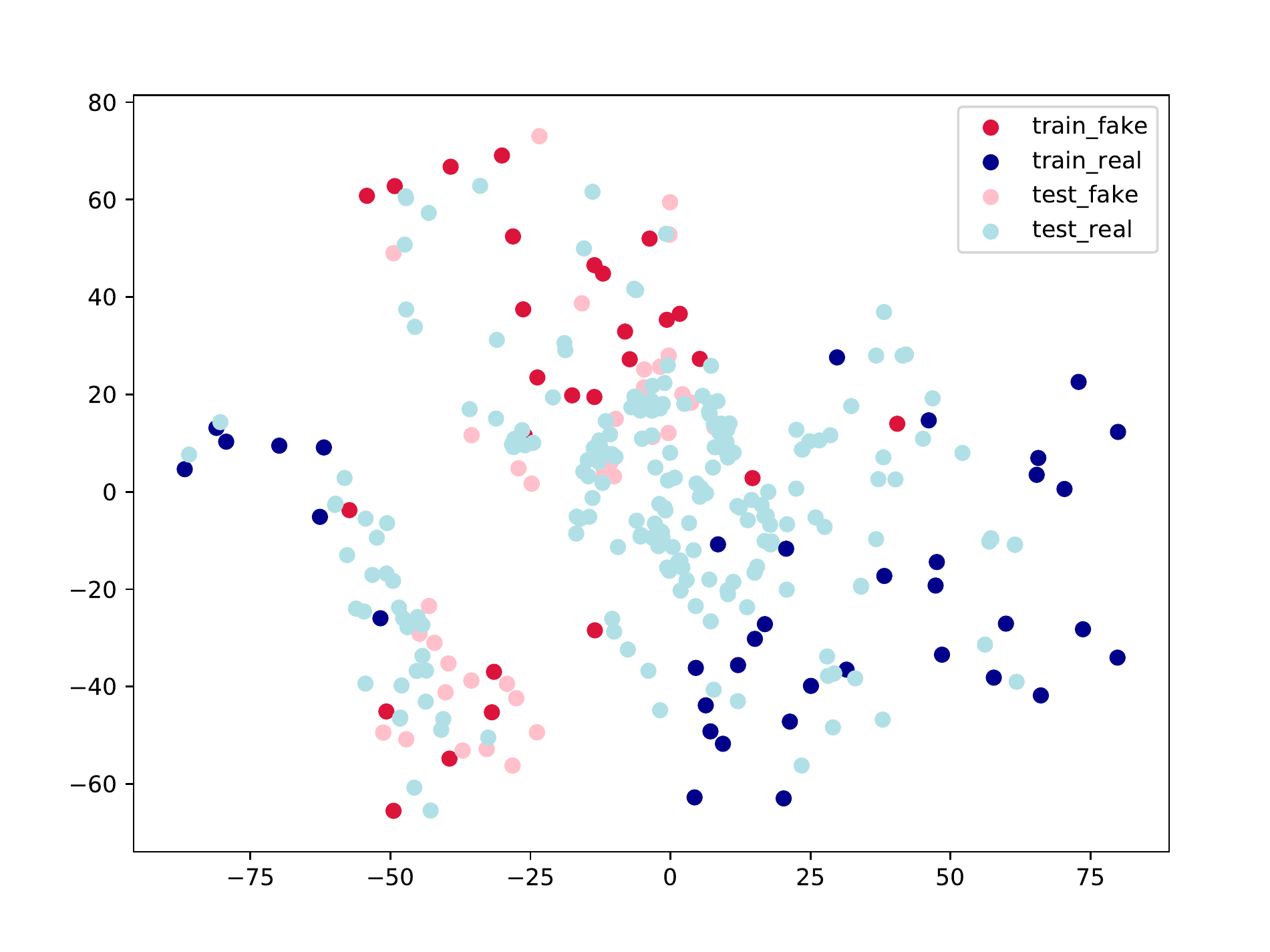}
         \caption{\yejindata~test set.}
         \label{fig:train_test_c99}
    \end{subfigure}
    \caption{Datasets distribution comparison with top 1\% influential training samples using t-SNE. Top 1\% influential samples are distributed fairly evenly over the whole training set (a), thus the extracted test features remain separable (b), and the \yejindata~distribution is captured better than trained with the full training set (c).}
    \label{fig:tsne2}
\end{figure}

Although both data set built to address COVID-19 fake-news and share the same data collection source, tweets, the results show that the models trained on \workshopdata~could achieve relatively lower performance on \yejindata~test set. (Note that the \yejindata~consists of the only test set with relatively smaller scale compared to \workshopdata.) For further understanding, we visualize features extracted by the best performing model right before the classification layers with t-SNE. As shown in Figure~\ref{fig:tsne1}, even though the features of \workshopdata~test set can distinguish the ``Fake'' and ``Real'' labels, the features of \yejindata~cannot separate the two labels quite well.

\subsection{How did smaller data help for generalization ability of the model?}
As mentioned in Subsection \ref{subsection:zero_shot_result}, higher cleansing percentage tends to lead to higher evaluation F1 score.
By using the model trained with top 1\% influential instances, we extract sentence representation as depicted in Figure~\ref{fig:tsne2}. Similar to in Figure~\ref{fig:tsne1}, the same number of instances from the test set are randomly selected for better understanding. Top 1\% influential instances are fairly evenly sampled from the whole training set, and this small subset of the training set is enough to produce the distribution to separate the test features, which supports the effectiveness of the influential score. Moreover, since the top 1\% samples are more sparse, the trained model can flexibly deal with samples from unseen distributions, resulting in extracted features of higher quality.

\begin{table}[t]
  \caption{Binary evaluation results of influence-based data cleansing model on \yejindata~test set. B-F1, B-Rec., and B-Pre.~denotes binary F1, binary recall, and binary precision scores respectively. Bold denotes the best performance over all experiments.}
  \label{table:result_binary}
  \centering
  \resizebox{\linewidth}{!}{
    \begin{tabular}{ccccccccc}
    \toprule
    Drop & \phantom{.} & \multicolumn{3}{c}{Fake}  & \phantom{.} & \multicolumn{3}{c}{Real} \\
    \cmidrule(lr){3-5} \cmidrule(lr){7-9}
    \% &  & B-F1 & B-Rec. & B-Pre. &  & B-F1 & B-Rec. & B-Pre. \\ 
    \midrule\midrule
    0\% & & 28.80 ± \scriptsize{1.06} & 99.29 ± \scriptsize{1.60} & 16.85 ± \scriptsize{0.71} & &	33.33 ± \scriptsize{5.25} & 20.12 ± \scriptsize{3.80} & 99.44 ± \scriptsize{1.24} \\
    1\% & & 29.06 ± \scriptsize{1.17} &  99.29 ± \scriptsize{1.60} &  17.03 ± \scriptsize{0.82} & & 34.46 ± \scriptsize{7.49} &  21.05 ± \scriptsize{5.43} &  99.58 ± \scriptsize{0.93} \\
    25\% & & 30.56 ± \scriptsize{1.23} &  99.29 ± \scriptsize{1.60} &  18.07 ± \scriptsize{0.88} & & 41.67 ± \scriptsize{6.11} &  26.51 ± \scriptsize{5.01} &  99.65 ± \scriptsize{0.78} \\
    50\% & & 31.02 ± \scriptsize{0.75} &  \textbf{100.0 ± \scriptsize{0.00}} & 18.36 ± \scriptsize{0.52} &  &43.16 ± \scriptsize{3.02} &  27.56 ± \scriptsize{2.49} &  \textbf{100.0 ± \scriptsize{0.00}} \\
    75\% & & 31.51 ± \scriptsize{0.85} &  99.29 ± \scriptsize{1.60} &  18.73 ± \scriptsize{0.66} & & 45.72 ± \scriptsize{4.47} &  29.77 ± \scriptsize{3.97} &  99.69 ± \scriptsize{0.70} \\
    99\% & & \textbf{37.17 ± \scriptsize{2.20}} &  81.43 ± \scriptsize{9.24} &  \textbf{24.28 ± \scriptsize{2.53}} & & \textbf{71.50 ± \scriptsize{6.92}} &  \textbf{57.79 ± \scriptsize{9.59}} &  95.23 ± \scriptsize{1.65}  \\ \bottomrule
    \end{tabular}
}
\end{table}

For the performance on \yejindata~test set, we take additional consideration on binary-Recall (B-Rec.), binary-Precision (B-Prec.), and binary-F1 (B-F1) scores to further analyze the generalization ability of the model. As shown in Table~\ref{table:result_binary}, the model with around 99\% data cleansing achieves the best per class F1-score with 37.17\% B-F1 score on the fake label and 71.50\% on the real label. In general, the ``Fake'' B-Pre and ``Real'' B-Rec scores increase as the cleansing percentage increase, while ``Real'' B-Pre and ``Fake'' B-Rec behave the other way around, which means the model with higher cleansing percentage capture more real news and reduce the number of false ``Fake'' label with the trade-off of capturing less true `Fake'' label. Overall, the B-F1 for each labels increases as the cleansing percentage increase. Our influence-based cleansing method outperforms the model without data cleansing by a large margin with  8.37\% for the ``Fake'' B-F1 and 38.17\% for the ``Real'' B-F1.

\section{Related Works}
\paragraph{COVID-19 Infodemic Research in Natural Language Processing}
In recent months, researchers took various approaches to tackle the problem of COVID-19 Infodemic. Wang et.~al~\cite{Wang2020CORD19TC} released centralized data  CORD-19 that covers 59,000 scholarly articles about COVID-19 and other related coronaviruses to encourage other studies. Singh et.~al~\cite{singh2020first} analyzed the global trend of tweets at the first emergence of COVID-19. To understand the diffusion of information, \cite{cinelli2020covid,shahi2020exploratory} analyze the patterns of spreading COVID-19 related information and also quantify the rumor amplification across different social media platforms. Alam et.~al~\cite{alam2020fighting} focuses on fine-grained disinformation analysis on both English and Arabic tweets for the interests of multiple stakeholders such as journalists, fact-checkers, and policymakers. Kar et.~al~\cite{kar2020rumours} proposes a multilingual approach to detect fake news about COVID-19 from Twitter posts.

\paragraph{Generalization ability of models}
As described in the previous section, several NLP studies involve emerging COVID-19 infodemic yet the generalization aspect is neglected although it is essential to accelerate industrial application development. In recent years, along with the introduction of numerous tasks in various domains, the importance of model generalization ability with a tiny amount or even without additional training datasets has been intensely discussed.
In general, recent works on model generalizability can be divided into two different directions: 1) adaptive training and 2) robust loss function.
In adaptive training, different meta-learning~\cite{finn2017maml} and fast adaptation~\cite{madotto2020adapterbot,winata2020fastadapt,liu2020crossner} approaches have been developed and show promising result for improving the generalization of the model over different domains. Another meta-learning approach, called meta transfer learning~\cite{winata2020mtl}, improves the generalization ability for a low-resource domain by leveraging a high-resource domain dataset.
In robust loss function, different kind of robust loss functions such as symmetric cross-entropy~\cite{wang2019symmetric}, generalized cross-entropy~\cite{zhang2018generalized}, and curriculum loss~\cite{lyu2019curriculum} have been shown to produce a more generalized model compared to cross-entropy loss due to its robustness towards noisy-labeled instances or so-called outliers from the training data. In addition to these approaches, data de-noising could actually improve model performance~\cite{lee2019team}, thus, a data cleansing technique with identifying influential instances in the training dataset is proposed to further improve the evaluation performance and generalization ability of the models~\cite{hara2019data,kobayashi-etal-2020-efficient}.

\section{Conclusion}
We investigated the COVID-19 fake-news detection task with an aim of achieving a robust model that could perform high for the CONSTRAINT shared task and also have high generalization ability with two separate approaches. The robust loss functions, compared to the traditional cross-entropy loss function, do not help much in improving F1-score on \workshopdata~but showed better generalization ability on \yejindata~with a fair trade-off as shown with the result comparison between RoBERTa-large with CE and CL. By performing influence data cleansing with high cleansing percentage ($\geq$ 25\%), we can achieve a better F1-score over multiple test sets. Our best model with 99\% cleansing percentage can achieve the best evaluation performance on \yejindata~with 61.10\% accuracy score and 54.33\% W-F1 score while still maintaining high enough test performance on \workshopdata. This suggests how we could use the labeled data to solve the problem of fake-news detection while model generalization ability should also be taken into account. For future work, we would like to combine the adaptive training, robust loss function with the influence score data cleansing method such that the resulting influence score can be made more robust for handling unseen or noisy data.
\bibliography{constraint-covid}
\bibliographystyle{splncs04}

\end{document}